\documentclass[conference]{IEEEtran}
\usepackage{times}
\usepackage{amsmath,amssymb,amsopn,amstext,amsfonts, amsthm}
\usepackage[ruled,vlined,linesnumbered]{algorithm2e}
\usepackage{algpseudocode}
\usepackage{graphicx}
\usepackage{booktabs}
\usepackage{graphicx}
\usepackage{subcaption}
\usepackage{multirow}
\usepackage[table,xcdraw]{xcolor}
\usepackage[numbers,sort&compress]{natbib}
\usepackage{fancyhdr}

\usepackage[numbers]{natbib}
\usepackage{multicol}
\usepackage[bookmarks=true]{hyperref}

\newcommand{\tp}{^{\mathrm{T}}}

\newcommand{\rBrac}[1]{\left({#1}\right)}

\newcommand{\Norm}[1]{\left\Vert{#1}\right\Vert}

\graphicspath{{figures/}}
\DeclareGraphicsExtensions{.png,.jpg,.eps,.pdf}

\begin{document}
	
	
	\title{Decentralized Spatial-Temporal \\Trajectory Planning for Multicopter Swarms}
	
	\author{Xin Zhou, Zhepei Wang, Xiangyong Wen, Jiangchao Zhu, Chao Xu and Fei Gao$^*$}
	


	%

	\maketitle
	
	\thispagestyle{fancy}     
	\fancyhf{}
	\fancyhead[L]{[Technical Report] FAST Lab, Zhejiang University}
	
	\pagestyle{fancy}            
	\fancyhf{}
	\fancyhead[L]{[Technical Report] FAST Lab, Zhejiang University}
	
	\let\thefootnote\relax\footnotetext{\scriptsize{*}Corresponding author.
		
		 All authors are with the State Key Laboratory of Industrial Control Technology, Institute of Cyber-Systems and Control, Zhejiang University, Hangzhou 310027, China and Huzhou Institute, Zhejiang University, Huzhou 313000, China. (email: {\tt $\{$iszhouxin, fgaoaa$\}$@zju.edu.cn}).}
	
	\begin{abstract}
		Multicopter swarms with decentralized structure possess the nature of flexibility and robustness, while efficient spatial-temporal trajectory planning still remains a challenge.
		This report introduces decentralized spatial-temporal trajectory planning, which puts a well-formed trajectory representation named MINCO into multi-agent scenarios.
		Our method ensures high-quality local planning for each agent subject to any constraint from either the coordination of the swarm or safety requirements in cluttered environments.
		Then, the local trajectory generation is formulated as an unconstrained optimization problem that is efficiently solved in milliseconds.
		Moreover, a decentralized asynchronous mechanism is designed to trigger the local planning for each agent.
		A systematic solution is presented with detailed descriptions of careful engineering considerations.
		Extensive benchmarks and indoor/outdoor experiments validate its wide applicability and high quality.
		Our software will be released for the reference of the community.
	\end{abstract}
	
	\IEEEpeerreviewmaketitle
	
	\section{Introduction}
	Aerial swarm robotics capable of three-dimensional operations show superior flexibility  (adaptability, scalability, and maintainability) and robustness (reliability, survivability, and fault-tolerance) compared to ground vehicles or single-agent systems \cite{chung2018survey}.
	In recent years, great amount of attentions have been paid to develop advanced architectures and algorithms that push the boundary of fully autonomous aerial swarms.
	
\begin{figure}[t]
		\centering
		\includegraphics[width=0.8\linewidth]{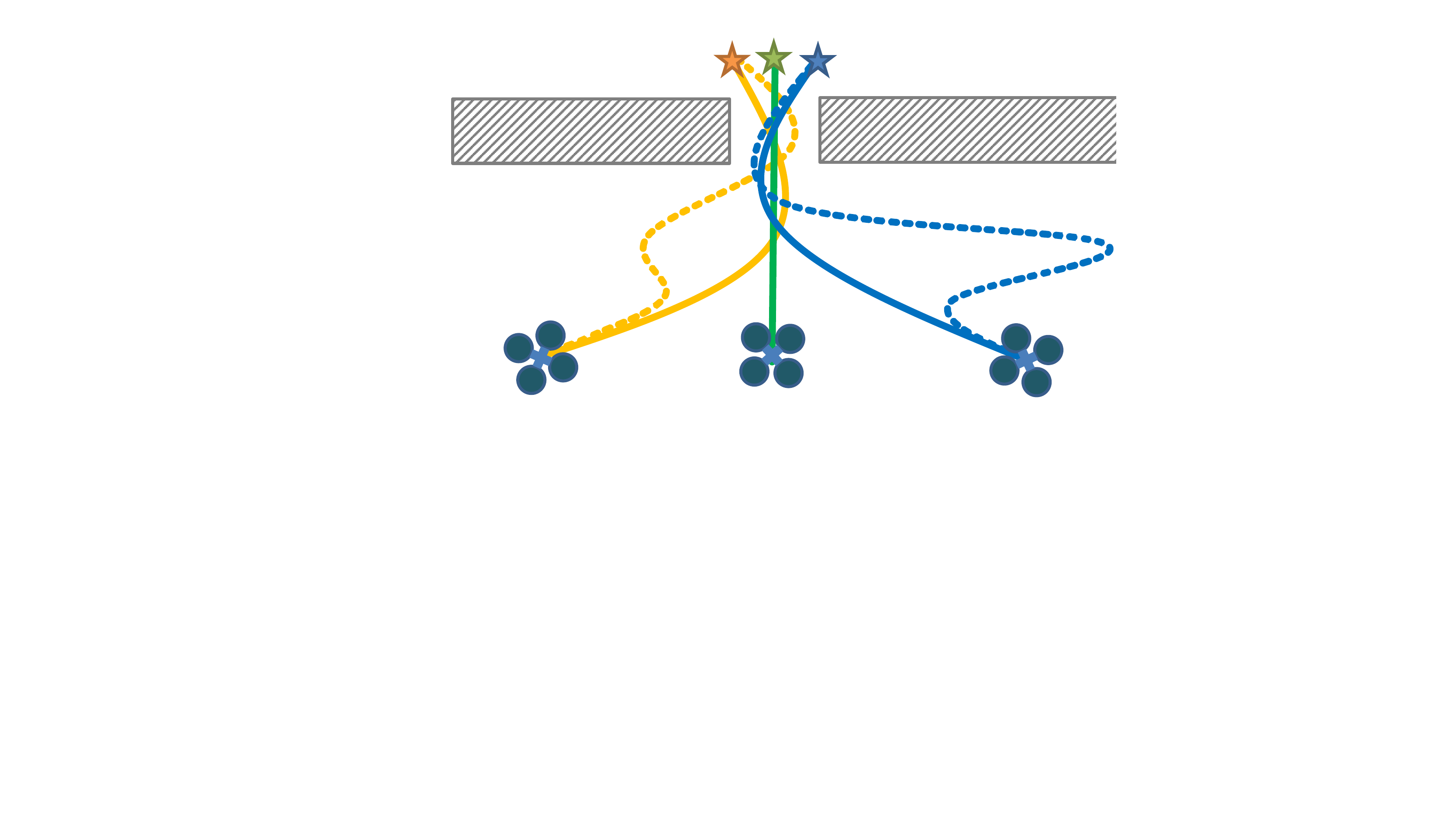}
		\captionsetup{font={small}}
		\caption{ Trajectory comparison with/without temporal optimization. Dotted curves:  without temporal optimization, agents use detours to delay time. Solid curves: with temporal optimization, shorter and smoother trajectories are generated.}
		\label{pic:bad_traj}
		\vspace{-0.3cm}
	\end{figure}

	The key module in the swarm is its planning algorithm, which dominates the efficiency and feasibility of the formation flight.
	Investigating the most underlying demand for swarm planning, it is expected to not only deform the shape of trajectories to avoid collisions, but also adjust the time profiles to exploit the sequential solution space and squeeze the feasibility of agents.
	Spatial-temporal joint trajectory optimization is primary to achieve this, but is challenging even for a single agent.
	If only spatial deformation is performed \cite{zhou2020egoswarm}, as shown in Fig.~\ref{pic:bad_traj}, agents tend to circumnavigate to wait for others while passing through a narrow passage, which hinders latter agents and results in inferior solutions.
	To this end, we adopt a recently developed trajectory representation named MINCO by \citet{Wang2021GCOPTER}, which is specially designed for spatial-temporal trajectory optimization for integrator chain systems.
	Moreover, since trajectory representations are critical but often ambiguous, starting with MINCO, we present a discussion (Sec.~\ref{sec:parameterization}) of several commonly used trajectory parameterization methods to developers in robotics community.
	
	\begin{figure}[t]
		\centering
		\includegraphics[width=1.0\linewidth]{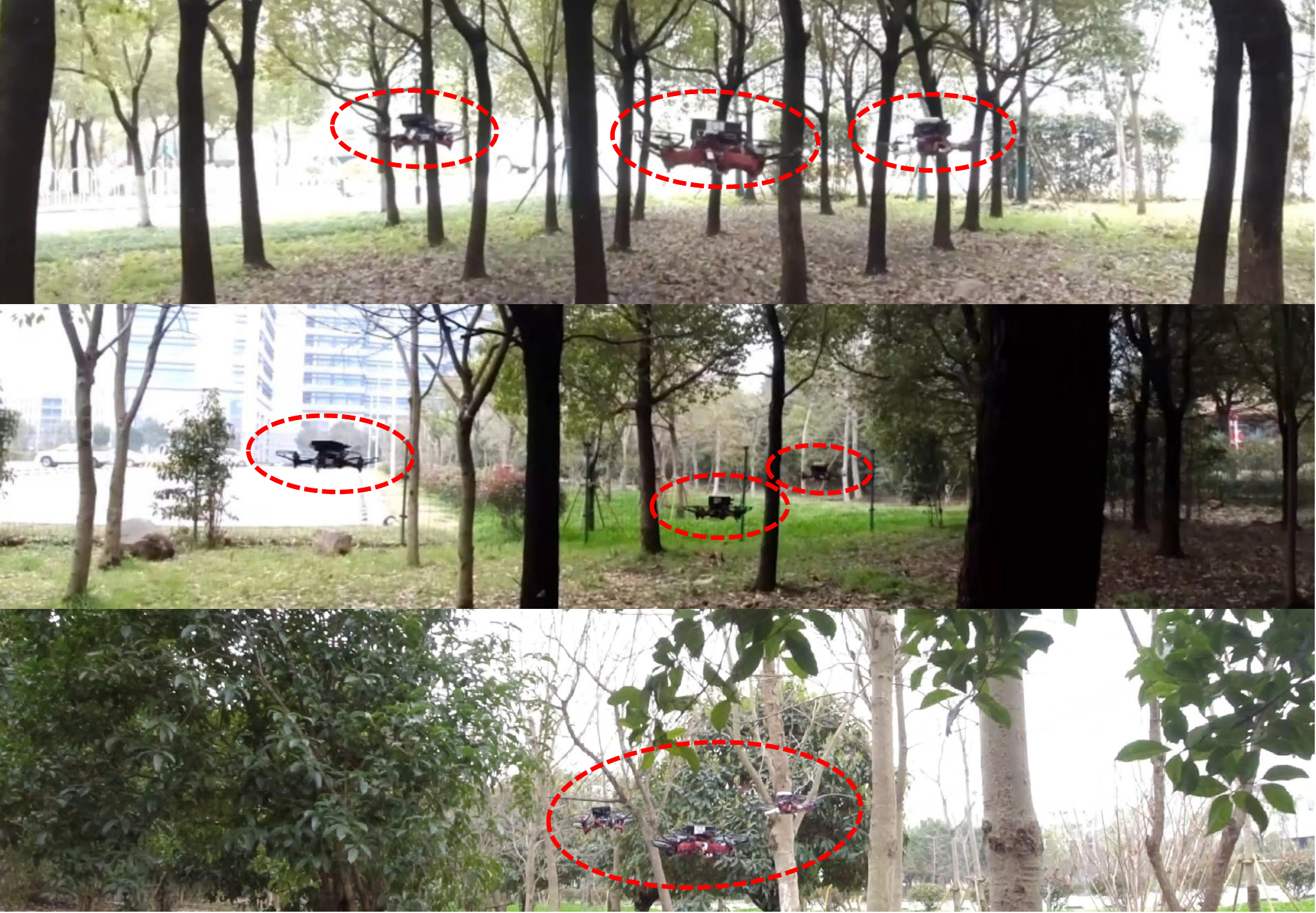}
		\captionsetup{font={small}}
		\caption{ Various outdoor experiments. Please watch the attached video for more information.}
		\label{pic:head}
	\end{figure}
	
	\begin{figure}[t]
		\centering
		\includegraphics[width=1.0\linewidth]{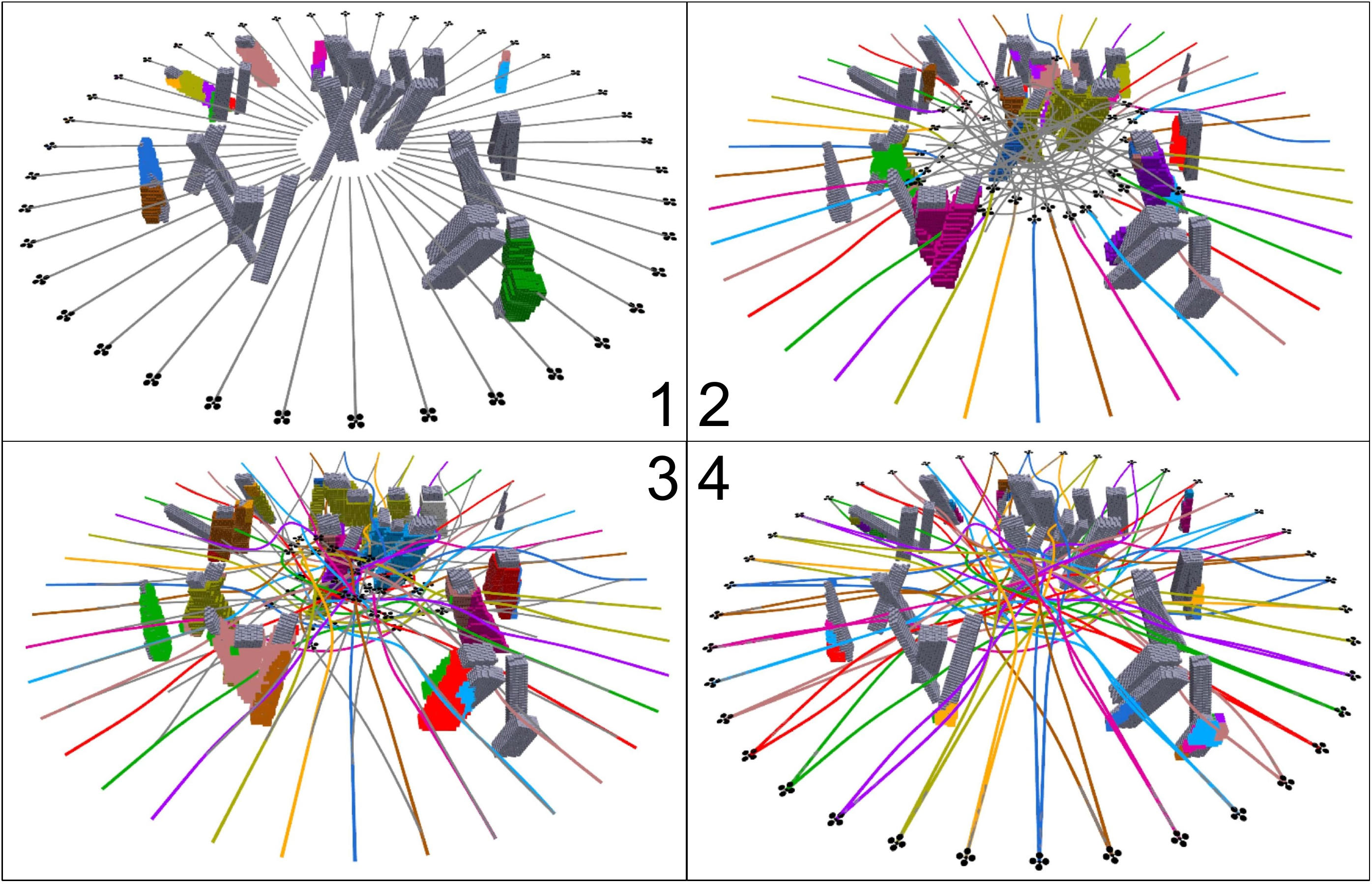}
		\captionsetup{font={small}}
		\caption{ Large scale position exchange. Colored curves indicate the positions agents have passed; gray curves are planned trajectories, colored obstacles represent the maps agents build. All the agents perform independent sensing, planning, and controlling. The program of all agents runs on a desktop CPU in real-time. }
		\label{pic:circle_sim}
	\end{figure}
	
	Based on MINCO, we propose a decentralized planner capable of spatial-temporal optimization for aerial swarms.
	To utilize this novel trajectory representation, several functions are designed to either penalize collision, restrict dynamical infeasibility or regularize the trajectory duration.
	Using the gradient propagation scheme of MINCO, the gradients of penalty functions on typical polynomial trajectories are efficiently propagated to those of MINCO.
	The planning problem is then formulated as an unconstrained optimization, which is solved efficiently with a customized solver in milliseconds.
	We further deploy our proposed algorithm into a hardware system built with careful engineering considerations.
	
	Besides algorithmic research, another fundamental requirement for the swarm is the architecture, which necessitates moderate inter-dependency, robust communication mechanism, and flexible system scale.
	This report presents a decentralized and asynchronously triggered planning strategy, which reduces the scale of the problem and distributes the computation burden to every single agent.
	Key components that bring this system to real-world, such as reciprocal drone detection, world frame alignments, networked trajectory broadcasting, and on-demand time synchronization, are also presented.
	Finally, the proposed fully autonomous swarm system is verified by extensive experiments in unknown, obstacle-rich, in/out-door environments as Fig. \ref{pic:head} and \ref{pic:circle_sim}.
	
	The main points of this report are summarized as follows:
	
	\begin{enumerate}
		\item A decentralized and asynchronous planning framework, which decouples the whole swarm planning problem and makes the system robust to communication failure.
		\item A spatial-temporal trajectory optimization method, extending MINCO representation to multi-agent scenarios.
		\item Engineering practices that bring the fully autonomous swarm to real-world. The software will be released after the reception of our previous work \cite{Wang2021GCOPTER} for the reference of the community.
	\end{enumerate}
	
	\begin{table*}[h]
		\parbox{.373\textwidth}{\caption{Pros and cons of trajectory parameterization methods in several frequently concerned aspects. More descriptions of these methods, including implementations, related papers, and detailed comparisons, are presented in Sec. \ref{sec:parameterization}.}
			\label{tab:traj_comp}}
		\renewcommand{\arraystretch}{1.35}
		\begin{tabular}{|c|c|c|c|c|c|}
			\hline
			\textbf{Method}       & \textbf{Continuity}                                                       & \textbf{Safety}          & \textbf{\begin{tabular}[c]{@{}c@{}}Dynamical\\ Feasibility\end{tabular}}          & \textbf{\begin{tabular}[c]{@{}c@{}}Temporal \\ Optimization\end{tabular}} & \textbf{\begin{tabular}[c]{@{}c@{}}Implementation\\ Difficulty\end{tabular}} \\ \hline
			\textbf{Polynomial}   & \multirow{2}{*}{\begin{tabular}[c]{@{}c@{}}Extra\\ Overhead\end{tabular}} & \multicolumn{2}{c|}{By Discretization}                                                                       & \multirow{2}{*}{Coupled}                                                  & Easy                                                                         \\ \cline{1-1} \cline{3-4} \cline{6-6} 
			\textbf{B\'ezier Curve} &                                                                           & \multicolumn{2}{c|}{\multirow{2}{*}{\begin{tabular}[c]{@{}c@{}}Analytical but \\ Conservative\end{tabular}}} &                                                                           & \multirow{2}{*}{Medium}                                                      \\ \cline{1-2} \cline{5-5}
			\textbf{B-Spline}     & \multirow{2}{*}{By Nature}                                                & \multicolumn{2}{c|}{}                                                                                        & Intractable                                                               &                                                                              \\ \cline{1-1} \cline{3-6} 
			\textbf{MINCO}        &                                                                           & \multicolumn{2}{c|}{By Discretization}                                                                       & Decoupled                                                                 & Difficult                                                                    \\ \hline
		\end{tabular}
	\end{table*}
	
	\section{Related Works}
	\label{sec:related_works}
	
	\subsection{Trajectory Parameterization for Aerial Robots}
	\label{sec:parameterization}
	Pros and cons of typical trajectory parameterization for aerial robots are summarized in Tab.~\ref{tab:traj_comp}.
	\textbf{Polynomial} curves are widely used by traditional works \cite{MelKum1105, richter2016polynomial, oleynikova2016continuous}, who mostly use polynomial coefficients as decision variables, along with explicit continuity constraints. 
	Many works \cite{richter2016polynomial, oleynikova2016continuous} deform a piecewise polynomial trajectory by adding intermediate waypoints.
	However, the distribution of waypoints requires a careful configuration to balance feasibility satisfaction and computational burden.
	For optimizing the time profile of a polynomial, the representative method uses gradient descent with numerical differences \cite{richter2016polynomial}.
	Nevertheless, this method requires a dense matrix inversion to transform waypoints to coefficients, making the optimization quickly intractable when the problem scales-up.
	\textbf{B\'ezier} and \textbf{B-Spline} share the \textit{convex hull} nature, which indicates that the curve is entirely contained in the convex hull of its control points \cite{tordesillas2019faster, gao2020teach}.
	Therefore, these curves are convenient to add constraints, which can be done by merely restricting the control points inside feasible convex regions.
	Many works \cite{park2020efficient, honig2018trajectory, tordesillas2020mader} follow this underlying property and show successful applications in aerial swarms.
	However, this property indeed brings conservativeness, as analyzed by \citet{tordesillas2020minvo}, preventing a trajectory from being aggressive near its physical limits.
	Besides, an $n$ order \textbf{B-Spline} is naturally $n-1$ order continuous \cite{de1978practical}. Thus no continuity constraint is required as long as the B-Spline has a qualified order. 
	For a \textbf{B\'ezier} curve, its time profile can be adjusted with acceptable complexity by its basis~\cite{gao2020teach}, while this is rather complicated for a B-Spline due to its time evaluation contains high nonlinearity.
	\textbf{MINCO} \cite{Wang2021GCOPTER} is a newly developed trajectory representation specialized for integrator chain systems. 
	The core feature of MINCO is that it efficiently handles a wide variant of constraints while retains spatial and temporal optimality.
	By discretizing the trajectory to add constraints, MINCO is less conservative but requires further checks after getting the solution.
	Although MINCO is the most complicated representation and requires a sophisticated implementation, it provides a solid fundamental for spatial-temporal drone swarm planning.
	
	\subsection{Decentralized Multicopter Swarms}
	
	For decentralized strategies, \textbf{VO} (velocity obstacle) is a lightweight approach to generate collision-free control commands \cite{van2011reciprocalnbody, van2011reciprocalcollision}.
	However, it does not produce trajectories with high-order continuity. Thus it is hard for a multicopter to execute.
	\citet{arul2020dcad} incorporate \textbf{VO} into an MPC as velocity constraints, which improves the trajectory quality significantly.
	Moreover, MPC is widely used in the literature on aerial swarm robotics.
	\citet{van2017distributed} achieve formation control for a group of vehicles.
	\citet{luis2019trajectory} use distributed MPC to perform point-to-point swarm transitions.
	Except that, \citet{chen2015decoupled} implements SCP for multiagent trajectory planning in non-convex scenarios by tightening collision constraints incrementally.
	However, the above methods require synchronization between the replans of different agents, which is hard to be guaranteed in real-world.
	To further decouple the system, \citet{liu2020mapper, tordesillas2020mader} propose decentralized and asynchronous planning strategies for drones to avoid static/dynamic obstacles and inter-vehicle collisions.
	However, the above optimization-based methods require at least tens of milliseconds, even several seconds, to compute a control command or a trajectory.
	What's more, those algorithms are only validated through simulations without integrating sensing, mapping, and localization.
	Early decentralized experimental results have been shown by \citet{mcguire2019minimal}. However, the naive minimum navigation method in that paper is doomed to produce discrete control commands with no consideration of system constraints.
	Recently, \citet{zhou2020egoswarm} builds up a fully autonomous multicpoter swarm system using B-Spline parameterized trajectories.
	Nevertheless, the difficulty of time adjustment of B-Spline curves results in a twisty trajectory shape when multiple agents need to pass through the same area.
	However, this limitation is broken in the proposed method.

	\section{Spatial-Temporal Trajectory Optimization For Swarms}
	\label{sec:trajgen}
	
	\subsection{Overview}
	\label{sec:overview}
	The capability of spatial-temporal optimization comes from a recently developed trajectory representation named $\mathfrak{T}_{\mathrm{MINCO}}$ \cite{Wang2021GCOPTER}.
	It's basic definition and features are explained in Sec.\ref{sec:minco}.
	Then Sec.\ref{sec:TimeIntCons} is a general description of how to add various continuous-time constraints to $\mathfrak{T}_{\mathrm{MINCO}}$ trajectories: by constraint quadrature.
	Finally in Sec.\ref{sec:problemFormu}, we formulate an unconstrained nonlinear optimization, which is then solved by gradient-decent methods. 
	To solve it in this way, costs and gradients of various objectives are defined and derived in Sec.\ref{sec:problemFormu} as well.
	
	\subsection{MINCO Trajectory Class}
	\label{sec:minco}
	
\begin{figure}[t]
		\centering
		\includegraphics[width=1.0\linewidth]{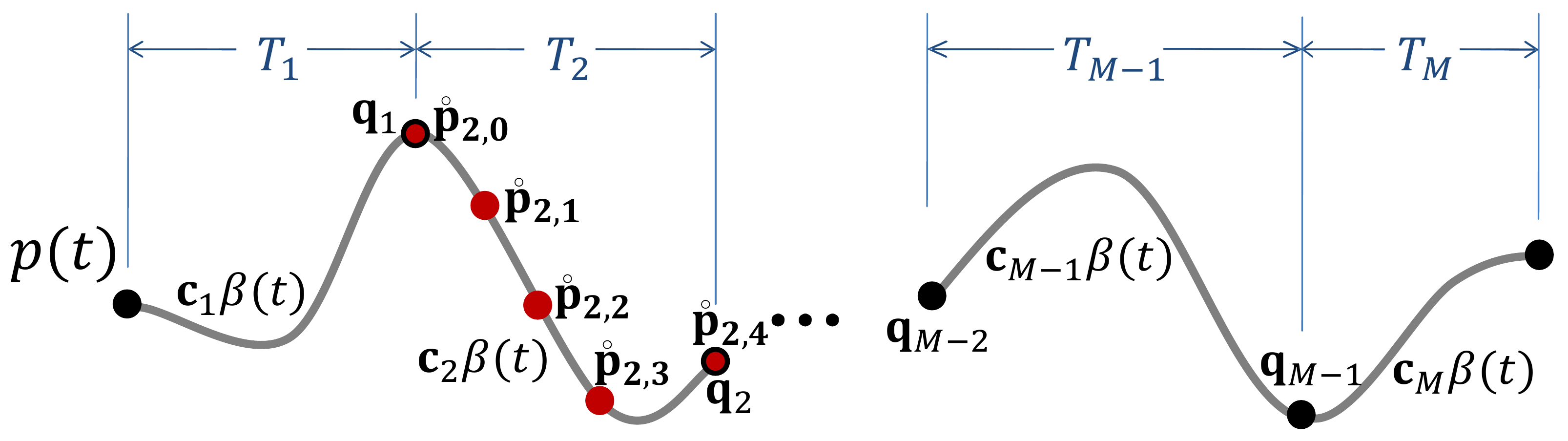}
		\captionsetup{font={small}}
		\caption{This figure illustrates the parameters of MINCO trajectory representation and its constraint evaluation. The trajectory is directly parameterized by every waypoint $\mathbf{q}_i$ and time $T_i$. Red dots represent \textit{constraint points} $\mathring{\mathbf{p}}$ (introduced in Sec. \ref{sec:TimeIntCons}), where any constraint is evaluated and propagated to every $\mathbf{q}_i$ and $T_i$. }
		\label{pic:minco}
	\end{figure}

	Thanks to the differential flatness of multicopters \cite{MelKum1105}, their motion planning is sufficient to be directly performed on a time differentiable curve.
	In this report, we adopt $\mathfrak{T}_{\mathrm{MINCO}}$ \cite{Wang2021GCOPTER} for our trajectory representation, which is a minimum control polynomial trajectory class defined as
	\begin{align*}
		\mathfrak{T}_{\mathrm{MINCO}} = \Big\{&p(t):[0, T]\mapsto\mathbb{R}^m \Big|~\mathbf{c}=\mathcal{M}(\mathbf{q},\mathbf{T}),~\\
		&~~\mathbf{q}\in\mathbb{R}^{m(M-1)},~\mathbf{T}\in\mathbb{R}_{>0}^M\Big\},
	\end{align*}
	where $p(t)$ is an $m$-dimensional $M$-piece polynomial trajectory with degree $N = 2s - 1$, $s$ the order of the relevant integrator chain, $\mathbf{c} = (\mathbf{c}_1\tp, \dots, \mathbf{c}_M\tp)\tp \in \mathbb{R}^{2Ms \times m}$ the polynomial coefficient, $\mathbf{q}=(\mathbf{q}_1,\dots,\mathbf{q}_{M-1})$ the intermediate waypoints and $\mathbf{T}=(T_1, T_2, \dots, T_M)\tp$ the time allocated for all pieces. 
	Specifically, the trajectory is evaluated as
	\begin{equation}
		\label{equ:polytraj}
		p(t)=p_i(t-t_{i-1}),~\forall t\in[t_{i-1},t_i].
	\end{equation}
	The $i$-th piece $p_i(t):[0, T_i]\mapsto\mathbb{R}^m$ is defined by
	\begin{equation}
		p_i(t)=\mathbf{c}_i\tp\beta(t),~\forall t\in[0,T_i],
	\end{equation}
	where $\beta(t):=[1, t, \cdots, t^N]\tp$ is the natural basis, $\mathbf{c}_i \in \mathbb{R}^{2s \times m}$ the coefficient matrix, $T_i=t_i-t_{i-1}$ and $T = \sum_{i=1}^{M} T_i$. The core of $\mathfrak{T}_{\mathrm{MINCO}}$ is the parameter mapping $\mathbf{c}=\mathcal{M}(\mathbf{q},\mathbf{T})$ constructed from Theorem 2 in \cite{Wang2021GCOPTER}. Technically, the mapping directly constructs a minimum control trajectory for an $m$-dimensional $s$-integrator chain for any specified initial and terminal conditions. An instance in $\mathfrak{T}_{\mathrm{MINCO}}$ is intuitively shown in Fig. \ref{pic:minco}.
	Since this report focuses on swarm planning, we only intuitively introduce the features of $\mathfrak{T}_{\mathrm{MINCO}}$.
	Detailed proofs are presented in \cite{Wang2021GCOPTER}.
	
	\textbf{Feature 1:} Compactly represented by $\mathbf{q}$ and $\mathbf{T}$, $\mathfrak{T}_{\mathrm{MINCO}}$ is a class of polynomial trajectories satisfying the following control effort minimization:
	\begin{subequations}
		\label{eq:MultistageMinimumControl}
		\begin{align}
			\min_{p(t)}&~{\int_{t_0}^{t_M}{p^{(s)}(t)\tp \mathbf{W}p^{(s)}(t)}\mathrm{d}t},\\
			\label{eq:ControlEffort}
			\mathit{s.t.}~&~p^{[s-1]}(t_0)=\bar{p}_o,~p^{[s-1]}(t_M)=\bar{p}_f,\\
			&~p(t_i)=q_i,~1\leq i<M,\\
			&~t_{i-1}<t_i,~1\leq i\leq M,
		\end{align}
	\end{subequations}
	where $\mathbf{W}\in\mathbb{R}^{m\times m}$ is a diagonal matrix with positive entries, $\bar{p}_o\in\mathbb{R}^{ms}$ and $\bar{p}_f\in\mathbb{R}^{ms}$ the specified initial and terminal conditions, $q_i\in\mathbb{R}^{m}$ is the given intermediate waypoints that the trajectory is enforced to pass at time $t_i$.
	
	\textbf{Feature 2:} The evaluation and differentiation of the mapping $\mathbf{c}=\mathcal{M}(\mathbf{q},\mathbf{T})$ enjoys linear complexity.
	A more specific correspondence can be expressed as the following function
	\begin{equation}
		\mathbf{M}(\mathbf{T})\mathbf{c}=\mathbf{b}(\mathbf{q}),
	\end{equation}
	where $\mathbf{M}(\mathbf{T})\in\mathbb{R}^{2Ms\times2Ms}$ is a banded matrix with nonsingularity for any $\mathbf{T}\succ\mathbf{0}$ ensured by Theorem 2 in \cite{Wang2021GCOPTER}, $\mathbf{b}(\mathbf{q})\in\mathbb{R}^{2Ms\times m}$.
	The conversion between these two trajectory representations, i.e., $(\mathbf{c}, \mathbf{T})$ and $(\mathbf{q}, \mathbf{T})$ is achieved using \textit{Banded PLU Factorization} with $O(M)$ linear time and space complexity. The evaluation is extremely fast, which takes about 1$\mu s$ per piece for minimum-jerk trajectory generation on a desktop CPU.
	
	\textbf{Feature 3:} Feature 2 allows any user-defined objective or penalty function $F(\mathbf{c},\mathbf{T})$ with available gradients applicable to $\mathfrak{T}_{\mathrm{MINCO}}$  parameterized in $(\mathbf{q},\mathbf{T})$. Specifically, the corresponding objective of $\mathfrak{T}_{\mathrm{MINCO}}$ is computed as
	\begin{equation}
		H(\mathbf{q},\mathbf{T})=F(\mathcal{M}(\mathbf{q},\mathbf{T}), \mathbf{T}).
	\end{equation}
	Then the mapping $\mathbf{c}=\mathcal{M}(\mathbf{q},\mathbf{T})$ gives a linear-complexity way to compute $\partial H/\partial\mathbf{q}$ and $\partial H/\partial\mathbf{T}$ from the corresponding $\partial F/\partial\mathbf{c}$ and $\partial F/\partial\mathbf{T}$.
	After that, a high-level optimizer is able to optimize the objective efficiently.
	
	\subsection{Time Integral Constraints}
	\label{sec:TimeIntCons}
	The way how various constraints are implemented closely relevant to trajectory parameterization methods.
	Conventionally, requirements for dynamical feasibility and collision avoidance are formulized as functional-type constraints $\mathcal{G}(p(t),\dots,p^{(s)}(t)) \preceq \mathbf{0}, \forall t\in[0,T]$.
	However, this formulation cannot be directly handled by constrained optimization, since $\mathcal{G}$ contains infinitely many inequality constraints.
	Thus, we transform $\mathcal{G}$ into a finite-dimensional one via integral of constraint violation.
	Practically, the integral is transformed into  weighted sum $J_\Sigma(\mathbf{c},\mathbf{T})$ of the sampled penalty function. In most cases that constraints are decoupled, i.e., constraints $\mathcal{G}(p^{[s]}(t))$ with $t_i \leq t < t_{i+1}$ are merely determined by $\mathbf{c}_i$ and $T_i$, the penalty for $i$-th trajectory piece is computed as
	\begin{equation}
		\label{eq:PieceTimeIntegralPenalty}
		J_i(\mathbf{c}_i,T_i,\kappa_i)=\frac{T_i}{\kappa_i}\sum_{j=0}^{\kappa_i}\bar{\omega}_j\chi\tp\max{(\mathcal{G}(\mathbf{c}_i,T_i,\frac{j}{\kappa_i}),\mathbf{0})^3},
	\end{equation}
	where $\kappa_i$ is the sample number on the $i$-th piece, $\chi\in\mathbb{R}^{n_g}_{\geq0}$ a vector of penalty weights with appropriately large entries, $(\bar{\omega}_0,\bar{\omega}_1,\dots,\bar{\omega}_{\kappa_i-1},\bar{\omega}_{\kappa_i})=(1/2,1,\cdots,1,1/2)$ are the quadrature coefficients following the trapezoidal rule~\cite{press2007numerical}.
	We define the points determined by $\{\mathbf{c}_i,T_i,{j}/{\kappa_i}\}$ as \textit{constraint points} $\mathring{\mathbf{p}}_{i,j} = p_i(({j}/{\kappa_i})T_i)$ with $p_i(t)$ the $i$-th polynomial piece.
	Then $\mathcal{G}(\mathbf{c}_i,T_i,{j}/{\kappa_i}) = \mathcal{G}(\mathring{\mathbf{p}}_{i,j})$.
	Note that $\mathring{\mathbf{p}}_{i,\kappa_i} = \mathring{\mathbf{p}}_{i+1,0}$.
	The cubic penalty is used here for its twice continuous differentiability.
	$J_\Sigma(\mathbf{c},\mathbf{T})$ is then defined as the summation:
	\begin{equation}
		\label{eq:TrajectoryTimeIntegralPenalty}
		J_\Sigma(\mathbf{c},\mathbf{T})=\sum_{i=1}^{M}{J_i(\mathbf{c}_i,T_i,\kappa_i)}.
	\end{equation}
	
	Since the gradients w.r.t. $(\mathbf{c}, \mathbf{T})$ are constructed from all $(\mathbf{c}_i,\mathbf{T}_i)$, we only give gradient templates to $\mathbf{c}_i$ and $T_i$ using the chain rule,
	\begin{equation}
		\label{eq:TimeIntGradC}
		\frac{\partial J_\Sigma}{\partial \mathbf{c}_i} = \frac{\partial J_\Sigma}{\partial \mathcal{G}} \frac{\partial \mathcal{G}}{\partial \mathbf{c}_i},
	\end{equation}
		\begin{equation}
		\label{eq:TimeIntGradT}
		\frac{\partial J_\Sigma}{\partial T_i} = \frac{J_i}{T_i} + \frac{\partial J_\Sigma}{\partial \mathcal{G}} \frac{\partial \mathcal{G}}{\partial t} \frac{\partial t}{\partial T_i},
	\end{equation}
	\begin{equation}
		\frac{\partial J_\Sigma}{\partial \mathcal{G}} = 3 \frac{T_i}{\kappa_i}\sum_{j=0}^{\kappa_i}\bar{\omega}_j\max{(\mathcal{G}(\mathbf{c}_i,T_i,\frac{j}{\kappa_i}),\mathbf{0})^2\circ\chi}.
	\end{equation}
	\begin{equation}
		\label{eq:pt_pTi}
		\frac{\partial t}{\partial T_i} = \frac{j}{\kappa_i}, ~~ t = \frac{j}{\kappa_i} T_i.
	\end{equation}
	From above equations, once the gradients of constraints $\mathcal{G}$ relative to polynomial coefficients $\mathbf{c}_i$ and time $t$ are evaluated, the gradients of $J_\Sigma(\mathbf{c},\mathbf{T})$ are efficiently computed.
	
	\subsection{Optimization Problem Construction}
	\label{sec:problemFormu}
	According to the above definitions of trajectory parameterization and constraints imposition, we present the complete optimization for trajectory planning in this section.
	The basic requirements on a desired trajectory are smoothness, dynamical feasibility, as well as safety among obstacles and other agents.
	Extra objectives such as minimization of control effort and execution time are also desired.
	In this report, we adopt $\mathfrak{T}_{\mathrm{MINCO}}$ to address above all concerns.
	Since $\mathfrak{T}_{\mathrm{MINCO}}$ is naturally guaranteed smooth by Theorem 2 in \cite{Wang2021GCOPTER}, no extra effort needs to be paid on trajectory continuity.
	
	In this work, we formulate the trajectory generation as an unconstrained optimization problem:
	\begin{equation}
		\label{eq:objfun}
		\min_{\mathbf{q},\mathbf{T}}~ \sum_{x}{\lambda_x J_x},
	\end{equation}
	where $J_x$ are $J_{\Sigma}$ instances, $\lambda_x$ is the weight for either an objective or penalty, subscripts $x=\{e,t,d,o,w,u\}$ represent control effort ($e$), execution time ($t$), dynamical feasibility ($d$), obstacle avoidance ($o$), swarm reciprocal avoidance ($w$), and uniform distribution of constraint points ($u$). Commonly, a sufficiently large weight suffices for penalties.
	The problem is solved via unconstrained nonlinear optimization.
	
	\subsubsection{Control Effort $J_e$}
	\label{sec:ControlEffort}
	Control effort follows the definition in Eq. \ref{eq:ControlEffort}.
	Without loss of generality, consider a single dimension of the $i$-th piece $p_i(t):[0, T_i]\mapsto\mathbb{R}$ of a polynomial spline, its derivatives with respect to $\mathbf{c}_i$ and $T_i$ are
	\begin{equation}
		\frac{\partial J_e}{\partial \mathbf{c}_i}=2 \left( \int_{0}^{T_i} \beta^{(s)}(t)\beta^{(s)}(t)^{\tp} dt \right) \mathbf{c}_i,
	\end{equation}
	\begin{equation}
		\frac{\partial J_e}{\partial T_i}=\mathbf{c}_i\tp\beta^{(s)}(T_i)\beta^{(s)}(T_i)^{\tp}\mathbf{c}_i.
	\end{equation}
	
	\subsubsection{Execution Time $J_t$}
	A shorter execution time is desirable, so we also minimize the weighted total execution time $J_t=\sum_{i=1}^{M} T_i$.
	Obviously, its gradient ${\partial J_t}/{\partial \mathbf{c}} = \mathbf{0}$, ${\partial J_t}/{\partial \mathbf{T}} = \mathbf{1}$.
	
	\subsubsection{Dynamical Feasibility Penalty $J_d$}
	\label{sec:DynamicalFeasibilityPenalty}
	For the trajectory generation of the multicopter, dynamical feasibility is always guaranteed by limiting the trajectory's derivatives.
	In our work, we limit the amplitude of velocity, acceleration, and jerk.
	Note that the dynamical feasibility penalty is acquired using time integral in Sec. \ref{sec:TimeIntCons}.
	Following the Eq. \ref{eq:PieceTimeIntegralPenalty}, constraints of velocity, acceleration, and jerk are denoted as
	\begin{equation}
		\label{eq:Jd_cost}
		\mathcal{G}_v = \dot{p}(t)^2 - v_m^2,~~~\mathcal{G}_a = \ddot{p}(t)^2 - a_m^2,~~~\mathcal{G}_j = \dddot{p}(t)^2 - j_m^2,
	\end{equation}
	where $v_m, a_m, j_m$ are maximum allowed velocity, acceleration and jerk.
	The corresponding gradients are
	\begin{equation}
		\label{eq:Jd_gradient}
		\frac{\partial \mathcal{G}_x}{\partial \mathbf{c}_i}=2{\beta^{(n)}(t)} {p^{(n)}}(t)\tp, ~~\frac{\partial \mathcal{G}_x}{\partial t}=2{\beta^{(n+1)}}(t)\tp \mathbf{c}_i p^{(n)}(t),
	\end{equation}
	where $x=\{v,a,j\}$, $n=\{1,2,3\}$, and $t=jT_i/\kappa_i+t_{i-1}, t \in [t_{i-1}, t_i]$, respectively.
	Substituting Eq. \ref{eq:Jd_cost} into \ref{eq:PieceTimeIntegralPenalty} and \ref{eq:Jd_gradient} into Eq.\ref{eq:TimeIntGradC}, \ref{eq:TimeIntGradT} gets the penalty and the gradient of $\mathbf{c}_i$ and $\mathbf{T}_i$;
	
	\subsubsection{Obstacle Avoidance Penalty $J_c$}
	\label{sec:ObstacleAvoidancePenalty}
\begin{figure}[t]
		\centering
		\includegraphics[width=0.9\linewidth]{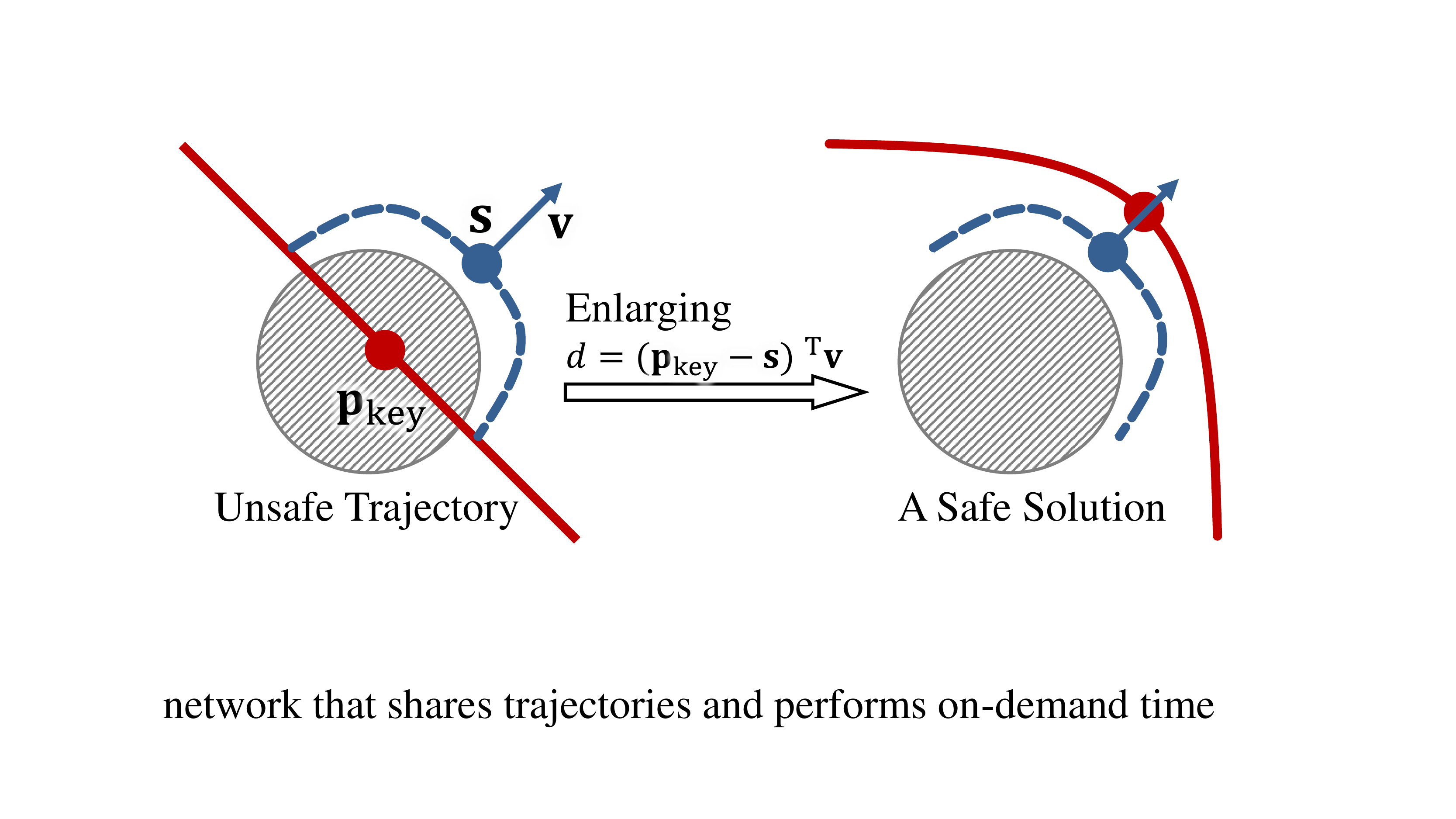}
		\captionsetup{font={small}}
		\caption{ Collision avoidance object formulation. The red curve is the optimized trajectory; the blue curve is a collision-free path starts and ends on the red curve. Then a fixed point $\mathbf{s}$ with a vector $\mathbf{v}$ is generated to determine the collision penalty. }
		\label{pic:ego}
		\vspace{-0.3cm}
	\end{figure}

	In this work, we adopt the front-end of collision evaluation from \citet{zhou2020ego}.
	Also a brief description is given here.
	In \cite{zhou2020ego}, the information for an unsafe trajectory to escape collision is extracted by comparing the unsafe initial trajectory with a collision-free guiding path.
	As depicted in Fig. \ref{pic:ego}, a fixed safe point $\mathbf{s}$ with a fixed safe vector $\mathbf{v}$ is recorded by a corresponding key point $\mathbf{p}_{\rm{key}}$ on the trajectory.
	Then the distance of $\mathbf{p}_{\rm{key}}$ to the obstacle which the $\{\mathbf{s}, \mathbf{v}\}$ pair generated is defined as
	\begin{equation}
		\label{eq:ObstalceDistance}
		d(\mathbf{p}_{\rm{key}},\{\mathbf{s}, \mathbf{v}\}) = (\mathbf{p}_{\rm{key}} - \mathbf{s})\tp\mathbf{v}.
	\end{equation}
	Using the distance information, the trajectory deforms iteratively during optimization.
	If $\mathbf{p}_{\rm{key}}$ discovers a new obstacle, a new $\{\mathbf{s}, \mathbf{v}\}$ pair is stacked to $\mathbf{p}_{\rm{key}}$'s records.
	Therefore each $\mathbf{p}_{\rm{key}}$ may record several $\{\mathbf{s}, \mathbf{v}\}$ pairs.
	
	In this work, constraint points $\mathring{\mathbf{p}}_{i,j}$ with $1\leq i \leq M, 0 < j \leq \kappa_i$ are selected as the key points $\mathbf{p}_{\rm{key}}$.
	To enforce safety, we penalize distance to obstacles less than a safe clearance $\mathcal{C}_o$.
	Therefore, obstacle avoidance constraint is defined as
	\begin{gather}
		\mathcal{G}_o(p(t))=\left(\cdots,\mathcal{G}_{o_k}(p(t)),\cdots\right)\tp\in\mathbb{R}^{N_{\mathbf{sv}}},\\
		\mathcal{G}_{o_k}(p(t)) = \mathcal{C}_o - d(p(t), \{\mathbf{s},\mathbf{v}\}_k),
	\end{gather}
	where $N_{\mathbf{sv}}$ is the number of $\{\mathbf{s},\mathbf{v}\}$ pairs recorded by a single $\mathring{\mathbf{p}}_{i,j}$ and $t=jT_i/\kappa_i$ the relative time on this piece.
	For the case that $\mathcal{C}_o \geq d(p(t), \{\mathbf{q},\mathbf{v}\}_k),0)$, the gradient is computed:
	\begin{equation}
		\frac{\partial \mathcal{G}_{o_k}}{\partial c_{i}} =  -\beta(t) \mathbf{v}\tp , ~~ \frac{\partial \mathcal{G}_{o_k}}{\partial t} =  -\mathbf{v}\tp\dot{p}(t).
	\end{equation}
	Otherwise, ${\partial \mathcal{G}_{o_k}}/{\partial c_{i}} = \mathbf{0}_{2s\times m}, {\partial \mathcal{G}_{o_k}}/{\partial t} = 0$.
	
	\subsubsection{Swarm Reciprocal Avoidance Penalty $J_w$}
	\label{sec:swarm_avoidance}
\begin{figure}[t]
		\centering
		\includegraphics[width=1.0\linewidth]{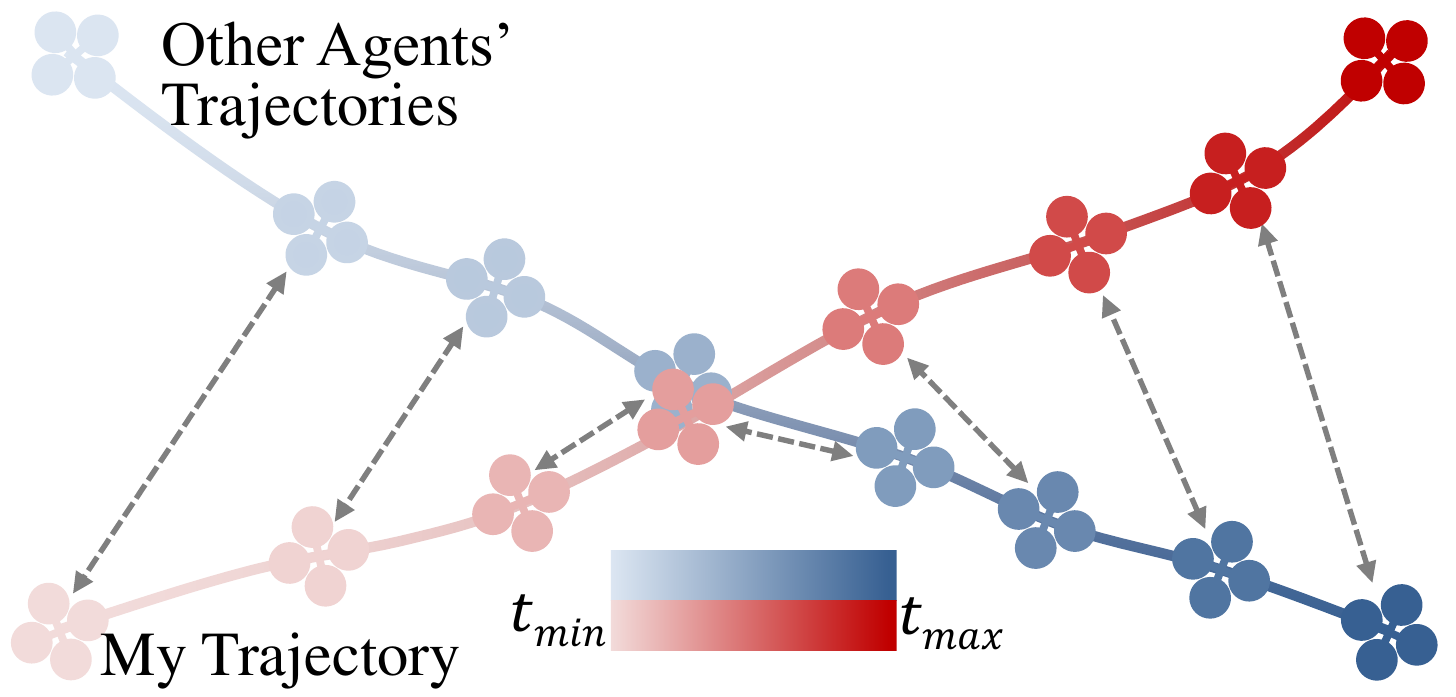}
		\captionsetup{font={small}}
		\caption{ Collision avoidance formulation between two agents. Gray dotted arrows indicate states at the identical global time. The goal of reciprocal avoidance is to maintain a safe distance at any time.}
		\label{pic:two_drone}
		\vspace{-0.5cm}
	\end{figure}

	In this work, a non-cooperative swarm framework is adopted, which means that one agent receives other agents' trajectories as constraints and generates trajectories only for itself.
	Considering the $u$-th agent in a multicpoter swarm containing $U$ agents, swarm collision avoidance is guaranteed when the trajectory $p_u(t)$ of agent $u$ maintains a distance greater than a safe clearance $\mathcal{C}_w$ to all the trajectories at the same global timestamps of other agents as depicted in Fig. \ref{pic:two_drone}. 
	However, one caution must be taken that we have always been using a relative time $t=jT_i/\kappa_i$ in our optimization, while the other agents' trajectories take an absolute timestamp $\tau=T_1 + \cdots + T_{i-1} + jT_i/\kappa_i$. 
	This indeed makes the penalty evaluated on the $i$-th piece depend on all its preceding piece times. Thus we denote by $G_{w}^{i,j}$ the constraint evaluated at the $j$-th constraint point on the $i$-th piece of $p_u(t)$.
	Therefore the reciprocal avoidance constraint is defined as
	\begin{gather}
		\mathcal{G}_w^{i,j}(p_u(t),\tau) = \left(\cdots, \mathcal{G}_{w_k}(p_u(t),\tau), \cdots\right)\tp\in\mathbb{R}^{U},\\
		\mathcal{G}_{w_k}^{i,j}(p_u(t),\tau) = \begin{cases} \mathcal{C}_w^2 - d^2(p_u(t), p_k(\tau)) & k\neq u,\\ 0 & k=u, \end{cases} \\
		d(p_u(t), p_k(\tau)) = \left\| \mathbf{E}^{1/2}(p_u(t) - p_k(\tau)) \right\|,
	\end{gather}
	where $p_k(\tau)$ is the trajectory of the $k$-th agent that the $u$-th agent has to avoid at the same but relative stamp $t$. The matrix $\mathbf{E}:=\rm{diag}(1,1,1/c)$ with $c>1$ transforms Euclidean distance into \textit{ellipsoidal distance} with the minor axes at the z-axis to relieve downwash risk from rotors.
	Squared distance is used to avoid \textit{square root} operations.
	
	When $\mathcal{C}_w^2 \geq d^2(p_u(t), p_k(\tau))$, the gradient to $\mathbf{c}_i$ is
	\begin{equation}
		\label{equ:G_wk_to_c}
		\frac{\partial \mathcal{G}_{w_k}^{i,j}}{\partial \mathbf{c}_{i}} = \begin{cases} -2\beta(t)(p_u(t) - p_k(\tau))\tp\mathbf{E} & k\neq u,\\ \mathbf{0} & k=u. \end{cases}
	\end{equation}
	Then substituting the sum of $\mathcal{G}_{w_k}^{i,j}$ into Eq.\ref{eq:TimeIntGradC} gets the gradient $\partial J_w / \partial \mathbf{c}_{i}$.
	However, the gradient to $\mathbf{T}$ is more complicated, since the gradient template equation \ref{eq:TimeIntGradT} does not hold here. Considering previous time profile, the proper gradient to the preceding time $T_l$ for any $1\leq l\leq i$ should be computed as
	
	\begin{align}
		\frac{\partial J_{w}}{\partial T_l}&=\sum_{k=1}^{U}\frac{\partial J_{w_k}}{\partial T_l}=\sum_{k=1}^{U}\sum_{i=1}^{M}\sum_{j=0}^{\kappa_i}\frac{\partial J_{w_k}^{i,j}}{\partial T_l},\\
		\frac{\partial J_{w_k}^{i,j}}{\partial T_l} &= \frac{J_{w_k}^{i,j}}{T_l}+\frac{\partial J_{w_k}^{i,j}}{\partial \mathcal{G}_{w_k}^{i,j}} \frac{\partial \mathcal{G}_{w_k}^{i,j}}{\partial T_l}, \\
		\frac{\partial \mathcal{G}_{w_k}^{i,j}}{\partial T_l}&=\frac{\partial \mathcal{G}_{w_k}^{i,j}}{\partial t}\frac{\partial t}{\partial T_l}+\frac{\partial \mathcal{G}_{w_k}^{i,j}}{\partial \tau}\frac{\partial \tau}{\partial T_l},\\
		\frac{\partial \mathcal{G}_{w_k}^{i,j}}{\partial t} &= \begin{cases} 2\left(p_k(t) - p_u(\tau)\right)\tp\mathbf{E}\dot{p}_u(t) & k\neq u,\\ 0 & k=u, \end{cases} \\
		\frac{\partial \mathcal{G}_{w_k}^{i,j}}{\partial \tau} &=  \begin{cases} 2\left(p_u(t) - p_k(\tau)\right)\tp\mathbf{E}\dot{p}_k(\tau) & k\neq u,\\ 0 & k=u, \end{cases} \\
		\frac{\partial t}{\partial T_l} =&\begin{cases} \frac{j}{\kappa_i} & l=i,\\ 0 & l<i, \end{cases}~~\frac{\partial \tau}{\partial T_l} = \begin{cases} \frac{j}{\kappa_i} & l=i,\\ 1 & l<i. \end{cases}
	\end{align}
	Note that $J_{w_k}^{i,j}$ denotes the swarm penalty of $j$-th constraint point at $i$-th polynomial piece to agent $k$'s current trajectory.
	The trajectory $p_k(\tau)$ for any other agent is already known, thus its derivatives are all available in computation. 
	
	\subsubsection{Uniform Distribution Penalty $J_v$}
	\label{sec:UniformDistributionPenalty}
\begin{figure}[t]
		\centering
		\includegraphics[width=0.8\linewidth]{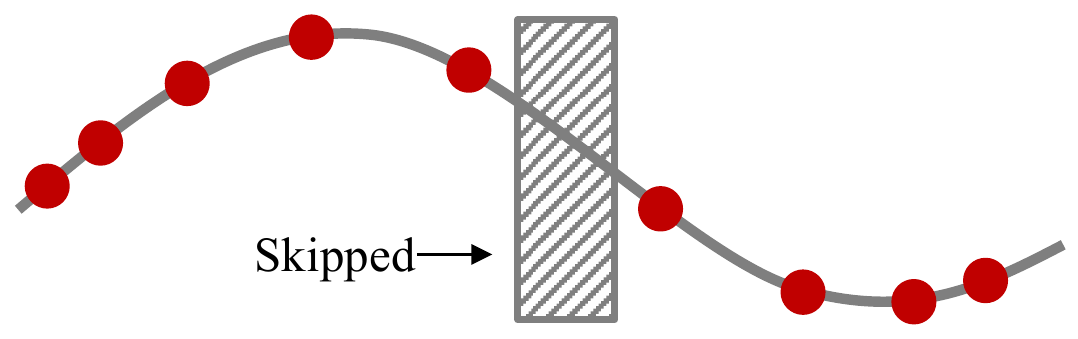}
		\captionsetup{font={small}}
		\caption{ Non-uniformly distributed constraint points (red dots) fail to detect a thin obstacle. }
		\label{pic:skipobs}
	\end{figure}

	The purpose of this penalty is to make the constraint points $\mathring{\mathbf{p}}_{i,j}$ equally spaced for all $1\leq i \leq M$ and $0 < j \leq \kappa_i$, which is a simpler substitution to the space-uniform variant of $\mathfrak{T}_{\mathrm{MINCO}}$ mentioned in \citep{Wang2021GCOPTER}.
	It is necessary since if all $\mathfrak{T}_{\mathrm{MINCO}}$ parameters $\mathbf{q}$ and $\mathbf{T}$ are optimized freely, some pieces of the trajectory tend to vanish to reduce the total cost.
	This phenomenon is harmful for trifold reasons. Firstly, $T_i = 0$ for the $i$-th trajectory piece is an undefined singular point for $\mathfrak{T}_{\mathrm{MINCO}}$, which does not consider consecutive aligned waypoints.
	Secondly, since the spatial collision avoidance is constrained by a finite number of constraint points,
	non-uniform distribution increases the possibility of skipping some tiny or thin obstacles, as shown in Fig. \ref{pic:skipobs}.
	Thirdly, since we compute the distance to obstacles following Eq. \ref{eq:ObstalceDistance}, the obstacles are modeled as planes.
	The accuracy of this modeling decreases when constraint points move along the plane.
	
	To enforce the uniform distribution of constraint points, we penalize the variance of the squared distances between each pair of adjacent constraint points. For simplicity, we denote $N_c=\sum_{i=0}^M{\kappa_i}$ as the total number of distances. The variance is computed as
	\begin{equation}
		J_u = \mathrm{E}\left[\mathbf{D}^2\right] - \mathrm{E}\left[\mathbf{D}\right]^2=\frac{1}{N_c}\Norm{\mathbf{D}}_2^2-\frac{1}{N_c^2}\Norm{\mathbf{D}}_1^2,
	\end{equation}
	where $\mathbf{D}\in\mathbb{R}^{N_c}$ is defined by 
	\begin{equation}
		\mathbf{D} = \rBrac{\left\|\mathring{\mathbf{p}}_{1,1} - \mathring{\mathbf{p}}_{1,0}\right\|^2_2, \cdots, \left\|\mathring{\mathbf{p}}_{M,\kappa_M} - \mathring{\mathbf{p}}_{M,\kappa_M-1}\right\|^2_2}.
	\end{equation}
	$\mathbf{D}$ is the squared distance vector for all $N_c+1$ constraint points. 
	Denote by $\mathbf{D}_k$ the $k$-th entry in $\mathbf{D}$ based on the conversion between two kinds of indices $k=j+\sum_{l=1}^{i-1}{\kappa_l}$. The gradient is computed as 
	\begin{align}
		\frac{\partial J_u}{\partial \mathbf{c}_i}&=\sum_{j=1}^{\kappa_i}\beta\rBrac{\frac{jT_i}{\kappa_i}}\rBrac{\frac{\partial J_u}{\partial \mathring{\mathbf{p}}_{i,j}}}\tp, \\
		\frac{\partial J_u}{\partial T_i}&=\frac{1}{\kappa_i}\sum_{j=1}^{\kappa_i}j\rBrac{\frac{\partial J_u}{\partial \mathring{\mathbf{p}}_{i,j}}}\tp\mathbf{c}_i\tp\dot{\beta}\rBrac{\frac{jT_i}{\kappa_i}},\\
		\frac{\partial J_u}{\partial \mathring{\mathbf{p}}_{i,j}}&=\frac{4}{N_c}\rBrac{\mathbf{D}_{k-1}-\frac{\Norm{\mathbf{D}}_1}{N_c}}\rBrac{\mathring{\mathbf{p}}_{i,j}-\mathring{\mathbf{p}}_{i,j-1}}\nonumber,\\
		&~~~-\frac{4}{N_c}\rBrac{\mathbf{D}_{k}-\frac{\Norm{\mathbf{D}}_1}{N_c}}\rBrac{\mathring{\mathbf{p}}_{i,j+1}-\mathring{\mathbf{p}}_{i,j}}.
	\end{align}
	
	\subsubsection{Temporal Constraint Elimination}
	An open-domain constraint is $\mathbf{T} \succ \mathbf{0}$.
	Unlike previous constraints from Sec.\ref{sec:DynamicalFeasibilityPenalty} to \ref{sec:UniformDistributionPenalty} which are transformed into penalty functions, this constraint is directly eliminated by a diffeomorphism map as is done in \cite{Wang2021GCOPTER}:
	\begin{equation}
		\label{eq:TDiffTransformation1}
		T_i=e^{\tau_i},
	\end{equation}
	where $\tau_i$ is the unconstrained virtual time. 
	Note that the meaning of notion $\tau$ here is different from Sec.\ref{sec:swarm_avoidance}.
	Gradients w.r.t. $T_i$ is propagated to $\tau_i$ following ${\partial J}/{\partial \tau_i} = ({\partial J}/{\partial T_i}) e^{\tau_i}$.
	More generally, any twice continuously differentiable bijection $f: \mathbb{R} \mapsto \mathbb{R}_{>0}$ suffices for this map.

	\begin{table*}[]
		\parbox{.276\textwidth}{\caption{Comparisons in an $8\times8m$ empty space containing eight agents with radius of $0.25m$. Maximum velocity and acceleration are set to $1.7m/s$ and $6m/s^2$. \textit{Safety Ratio} is the minimum agent interval divided by two times of radius. \textbf{int($a^2$)} and \textbf{int($j^2$)} are time integral of squared acceleration and jerk, indicating smoothness and control effort. The units of time and distance are seconds and meters.}
			\label{tab:no_obs}}
		\renewcommand{\arraystretch}{1.35}
		\begin{tabular}{|c|l|ccccccc|}
			\hline
			\textbf{\begin{tabular}[c]{@{}c@{}}Online/\\ Offline\end{tabular}} & \multicolumn{1}{c|}{\textbf{Method}} & \textbf{\begin{tabular}[c]{@{}c@{}}Solver \\ Time\end{tabular}} & \textbf{\begin{tabular}[c]{@{}c@{}}Traj.\\ Time\end{tabular}} & \textbf{\begin{tabular}[c]{@{}c@{}}Traj.\\ Length\end{tabular}} & \textbf{\begin{tabular}[c]{@{}c@{}}Safety\\ Ratio\end{tabular}} & \textbf{Safe? }                     & \textbf{int($a^2$)}    & \textbf{int($j^2$)}     \\ \hline
			& SCP, h\_scp=0.3s            & 1.46                                                   & 7.65                                                 & \underline{\textbf{9.64}}                                          & 0.267                                                  & {\color[HTML]{FE0000} No}  & 2.63          & 7.01           \\
			& SCP, h\_scp=0.17s           & 7.72                                                   & \underline{\textbf{7.40}}                                        & 9.70                                                   & 0.480                                                  & {\color[HTML]{FE0000} No}  & 3.14          & 17.5           \\ \cline{2-9}
			& RBP,no batches              & \textbf{0.320}                                         & 15.4                                                 & 11.3                                                   & 1.02                                                   & {\color[HTML]{007A1E} Yes} & \underline{\textbf{1.30}} & 0.608          \\
			\multirow{-4}{*}{Offline}                                 & RBP,batch\_size=4           & 0.413                                                  & 15.4                                                 & 11.5                                                   & 1.06                                          & {\color[HTML]{007A1E} Yes} & 1.32          & \underline{\textbf{0.604}} \\ \hline
			& Mader                       & 0.0239                                                 & \textbf{7.50}                                        & 12.2                                                   & 1.37                                                   & {\color[HTML]{007A1E} Yes} & 12.0          & 125.1         \\ \cline{2-9}
			& EGO, horizon=7.5m            & 0.000554                                               & 8.12                                                 & 10.1                                                   & 1.34                                                   & {\color[HTML]{007A1E} Yes} & 7.39          & 72.0           \\
			& EGO, horizon=10m             & 0.000844                                               & 8.17                                                 & 10.3                                                   & 1.62                                          & {\color[HTML]{007A1E} Yes} & 8.15          & 94.1           \\ \cline{2-9}
			& Proposed, $\kappa_i=5$          & \underline{\textbf{0.000465}}                                      & 7.57                                                 & \textbf{9.70}                                          & 1.17                                                   & {\color[HTML]{007A1E} Yes} & 5.33          & 30.4           \\
			\multirow{-5}{*}{Online}                                  & Proposed, $\kappa_i=8$           & 0.000557                                               & 7.58                                                 & 9.71                                                   & 1.22                                                   & {\color[HTML]{007A1E} Yes} & \textbf{5.15} & \textbf{28.2}  \\ \hline
		\end{tabular}
	\end{table*}

	\begin{table*}[]
		\parbox{.355\textwidth}{\caption{Comparisons in a $20\times$20$\times5m$ space with 100 obstacles and 8 agents. This table shares the same parameters, units, and notations as Tab .\ref{tab:no_obs}. }
			\label{tab:with_obs}}
		\renewcommand{\arraystretch}{1.35}
		\begin{tabular}{|c|cccccccc|}
			\hline
			\multicolumn{1}{|c|}{\textbf{Method}} & \textbf{\begin{tabular}[c]{@{}c@{}}Solver\\ Time\end{tabular}} & \textbf{\begin{tabular}[c]{@{}c@{}}Traj.\\ Time\end{tabular}} & \textbf{\begin{tabular}[c]{@{}c@{}}Traj.\\ Length\end{tabular}} & \textbf{\begin{tabular}[c]{@{}c@{}}Safety\\ Ratio\end{tabular}} & \textbf{\begin{tabular}[c]{@{}c@{}}Dist.\\ to Obs.\end{tabular}} & \textbf{Safe?}                      & \textbf{int($a^2$)} & \textbf{int($j^2$)} \\ \hline
			MADER                 & 0.0539                                              & 19.4                                                 & 30.25                                                 & 1.52                                                   & 0.466                                                  & {\color[HTML]{007A1E} Yes} & 39.6       & 960.9      \\ \hline
			EGO                 & 0.000882                                              & 19.6                                                 & 29.8                                                 & 1.57                                                   & 0.623                                                  & {\color[HTML]{007A1E} Yes} & 20.3       & 198.4      \\ \hline
			Proposed                     & \textbf{0.000755}                                              & \textbf{16.9}                                                 & \textbf{29.1}                                                  & 1.22                                                   & 0.600                                                  & {\color[HTML]{007A1E} Yes} & \textbf{13.5}       & \textbf{75.3}       \\ \hline
		\end{tabular}
	\vspace{0.5cm}
	\end{table*}
	
	

	\section{Benchmark}
	
	\begin{figure}[t]
		\centering
		\includegraphics[width=1.0\linewidth]{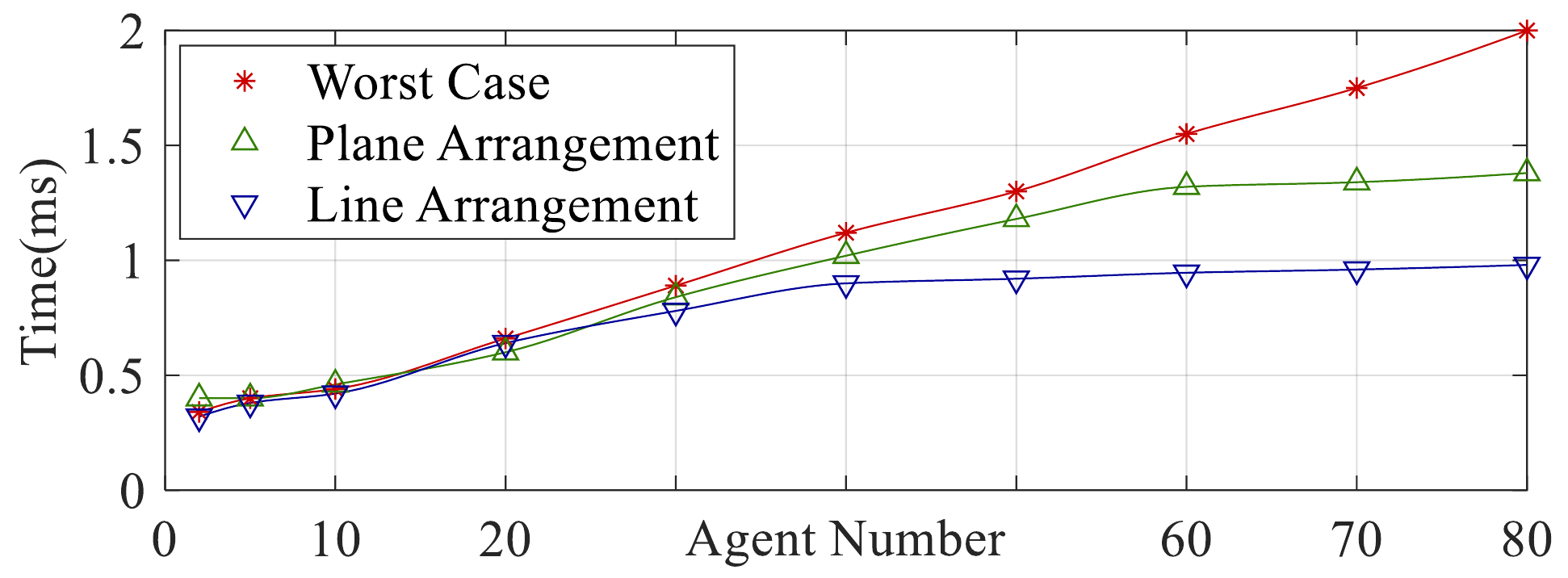}
		\captionsetup{font={small}}
		\caption{ Scalability evaluation in obstacle-free scenarios. Agents in the \textbf{\textit{Worst Case}} that considering all the others achieve linear complexity relative to the agent amount.
			If ignoring agents' trajectories outside the planning horizon, the complexity is further reduced. }
		\label{pic:scale}
	\end{figure}

	\begin{figure}[t]
		\centering
		\includegraphics[width=1.0\linewidth]{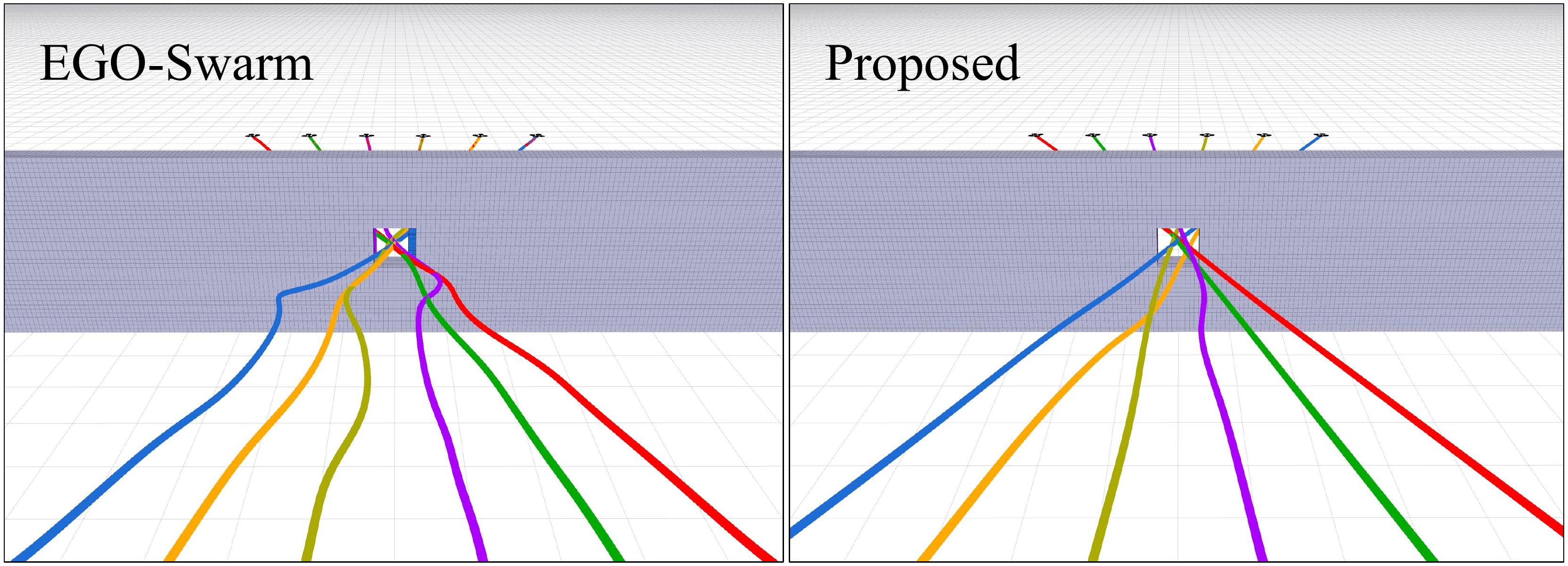}
		\captionsetup{font={small}}
		\caption{ Passing through a narrow gate with EGO-Swarm and the proposed planner. Due to spatial-temporal planning capability, the proposed method achieves smoother trajectories }
		\label{pic:egoswarm_bad}
	\end{figure}
	
	\subsection{Large Scale Simulation}
	We conduct tests on large-scale planning scenarios, where 40 agents perform position exchanging on a circle with a $12.5m$ radius, as depicted in Fig.\ref{pic:circle_sim}.
	
	\subsection{Scalability}
	Since a non-cooperative swarm framework is adopted, the complexity of the proposed method relative to agent number $U$ is $O(U)$.
	What's more, the complexity can be further reduced based on a consensus that the capacity of a given space is finite. 
	Therefore, the received trajectory is neglected if it is outside the planning horizon.
	Here we demonstrate the scalability on replanning time of each agent in three scenarios.
	In the \textbf{\textit{Worst Case}}, the trajectory ignoring mechanism is disabled.
	In the \textbf{\textit{Line Arrangement}} case, agents start in a straight line and fly to targets uniformly placed on another line $50m$ away.
	In the \textbf{\textit{Plane Arrangement}} case, agents are initialized on a plane with targets uniformly placed on another plane.
	The order of targets in all tests is randomly selected.
	Results are shown in Fig. \ref{pic:scale}
	
	\subsection{Comparisons}
	We compare the proposed planner against \textbf{SCP} \cite{augugliaro2012generation}, \textbf{RBP} \cite{park2020efficient}, \textbf{MADER} \cite{tordesillas2020mader}, and \textbf{EGO} (EGO-Swarm) \cite{zhou2020egoswarm}.
	All presented data is the average of 8 agents in 10 runs, except for \textbf{Safety Ratio} and \textbf{Distance to Obstacles} (in abbreviation \textbf{Dis. to Obs.}) are the minimal value in all records.
	
	\subsubsection{Comparison of Reciprocal Collision Avoidance \textbf{without} Static Obstacles}
	Results are given in Tab. \ref{tab:no_obs}.
	A bold term indicates a better value in the corresponding category (Online/Offline), while an underlined term indicates the best value among all planners. Indicated by Tab. \ref{tab:no_obs}, \textbf{SCP} achieves a better flight time at the sacrifice of trajectory safety. \textbf{RBP} achieves better dynamical performance $\mathbf{int}(a^2)$ and $\mathbf{int}(j^2)$ with the cost of a relatively long flight time. 
	Both of these methods fail to balance various aspects of trajectory quality. 
	Compared with offline methods, three online planners show more balances while requiring less computation by orders of magnitude. Among them, \textbf{MADER} shows more conservativeness due to the simplex representation of the trajectory, although the simplex enjoys almost the minimum volume.
	
	\subsubsection{Comparison of Reciprocal Collision Avoidance \textbf{with} Static Obstacles}
	Results are presented in Tab. \ref{tab:with_obs} for three online planners.
	Note that \textbf{MADER} uses a pre-built map while \textbf{EGO-Swarm} and the proposed method perform online sensing and mapping.
	From the data, \textbf{MADER} demands aggressive maneuvers for the vehicle while the other two are more superior in their trajectory smoothness. Another experiment to intuitively visualize the benefits of temporal optimization is illustrated in Fig. \ref{pic:egoswarm_bad}.
	
	\subsubsection{A Brief Conclusion}
	From the comparison, the proposed method achieves top-level performance with the shortest computation time.
	This achievement is attributed to low complexity of $\mathfrak{T}_{\rm{MINCO}}$-based problem  formulation and decoupling of spatial-temporal parameters.
	All the programmers run on an Intel Core i7-10700KF CPU at 5.1GHz.
	
	
	\begin{figure}[t]
		\centering
		\includegraphics[width=1.0\linewidth]{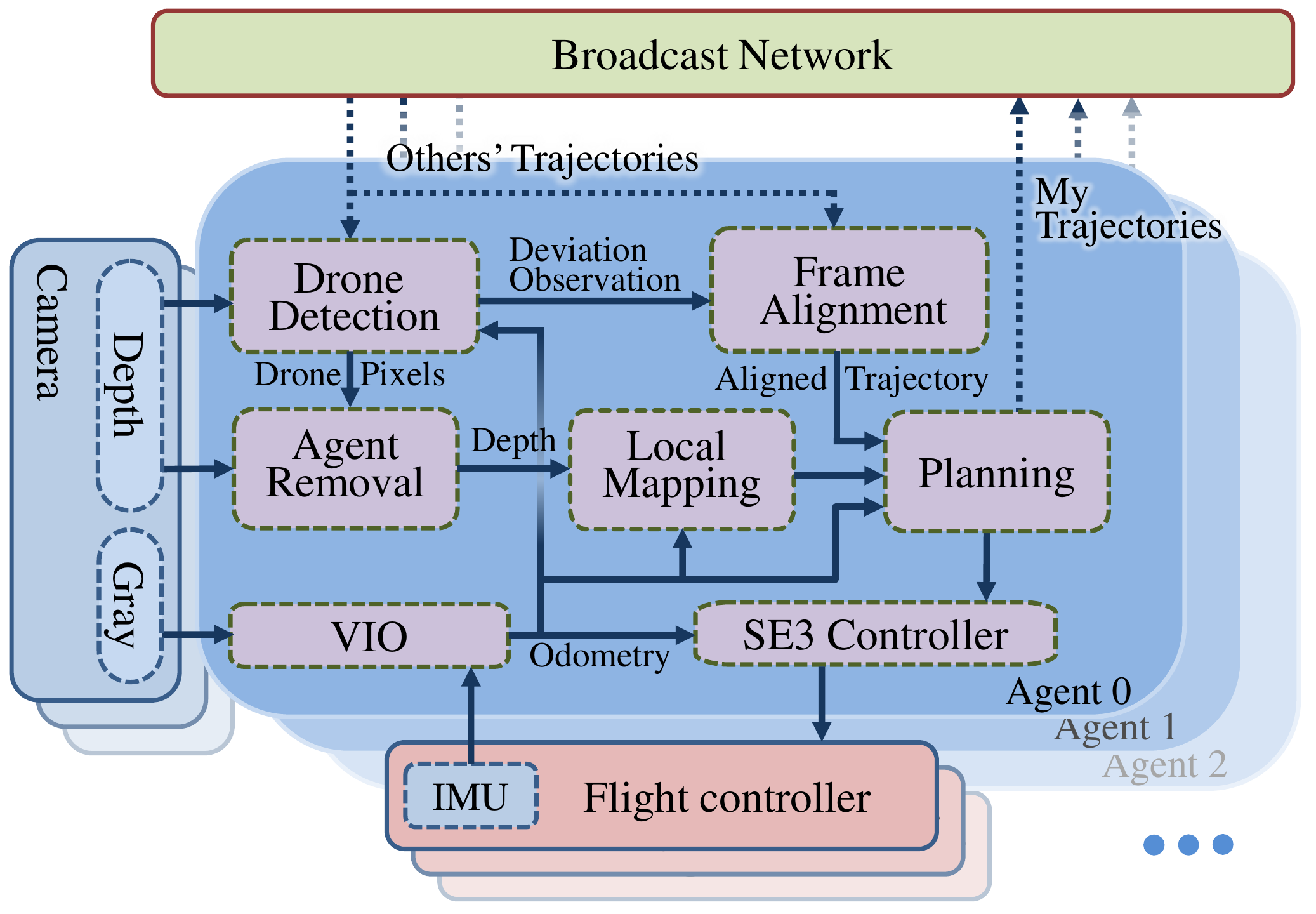}
		\captionsetup{font={small}}
		\caption{ System Architecture. }
		\label{pic:sys_archi}
		\vspace{-0.3cm}
	\end{figure}

	\begin{figure*}[t]
		\centering
		\begin{subfigure}{0.3\linewidth}
			\includegraphics[width=1\linewidth]{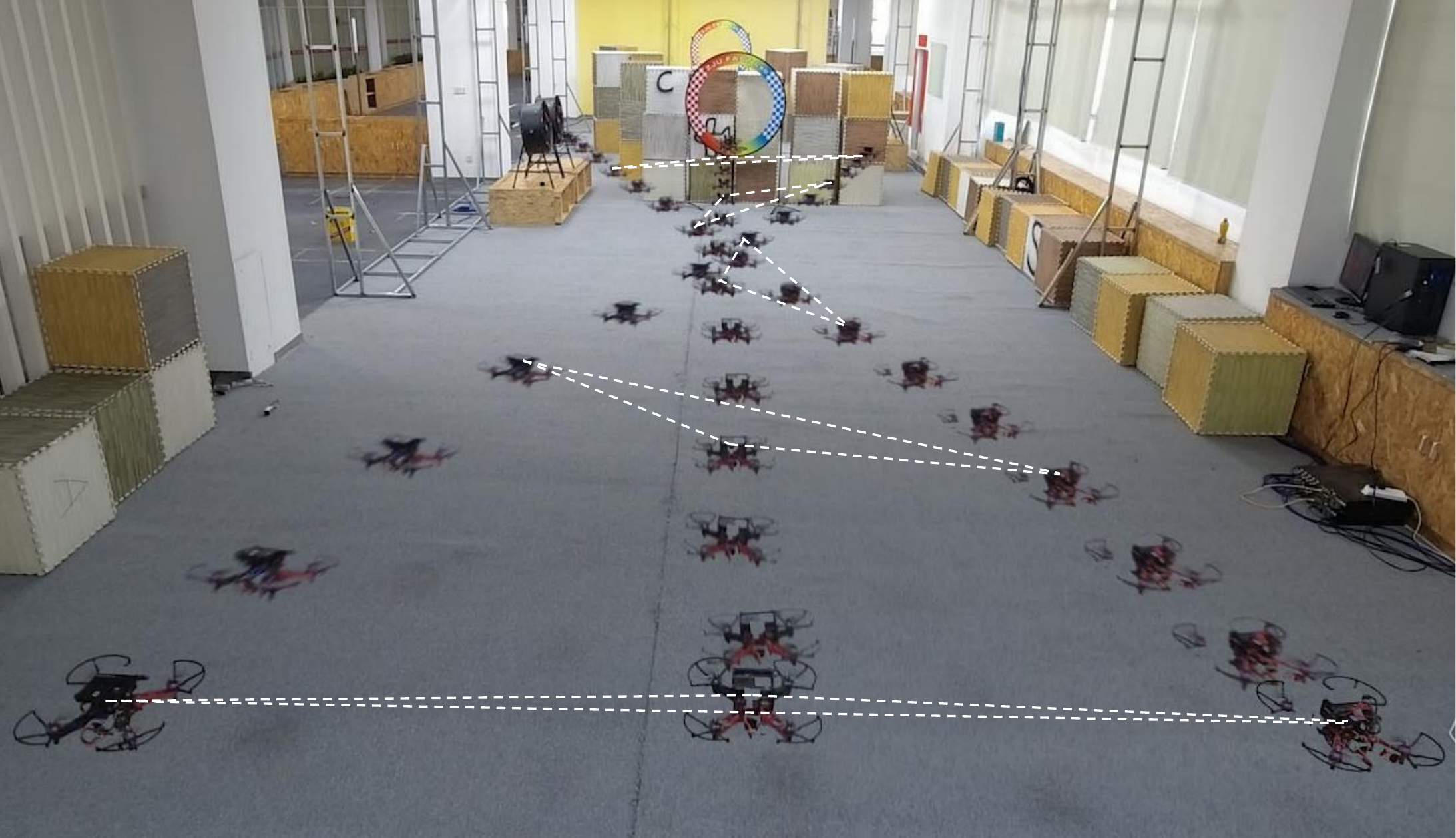}
			\captionsetup{font={small}}
			\caption{Empty Space}
			\label{pic:indoorempty}
		\end{subfigure}
		\begin{subfigure}{0.3\linewidth}
			\includegraphics[width=1\linewidth]{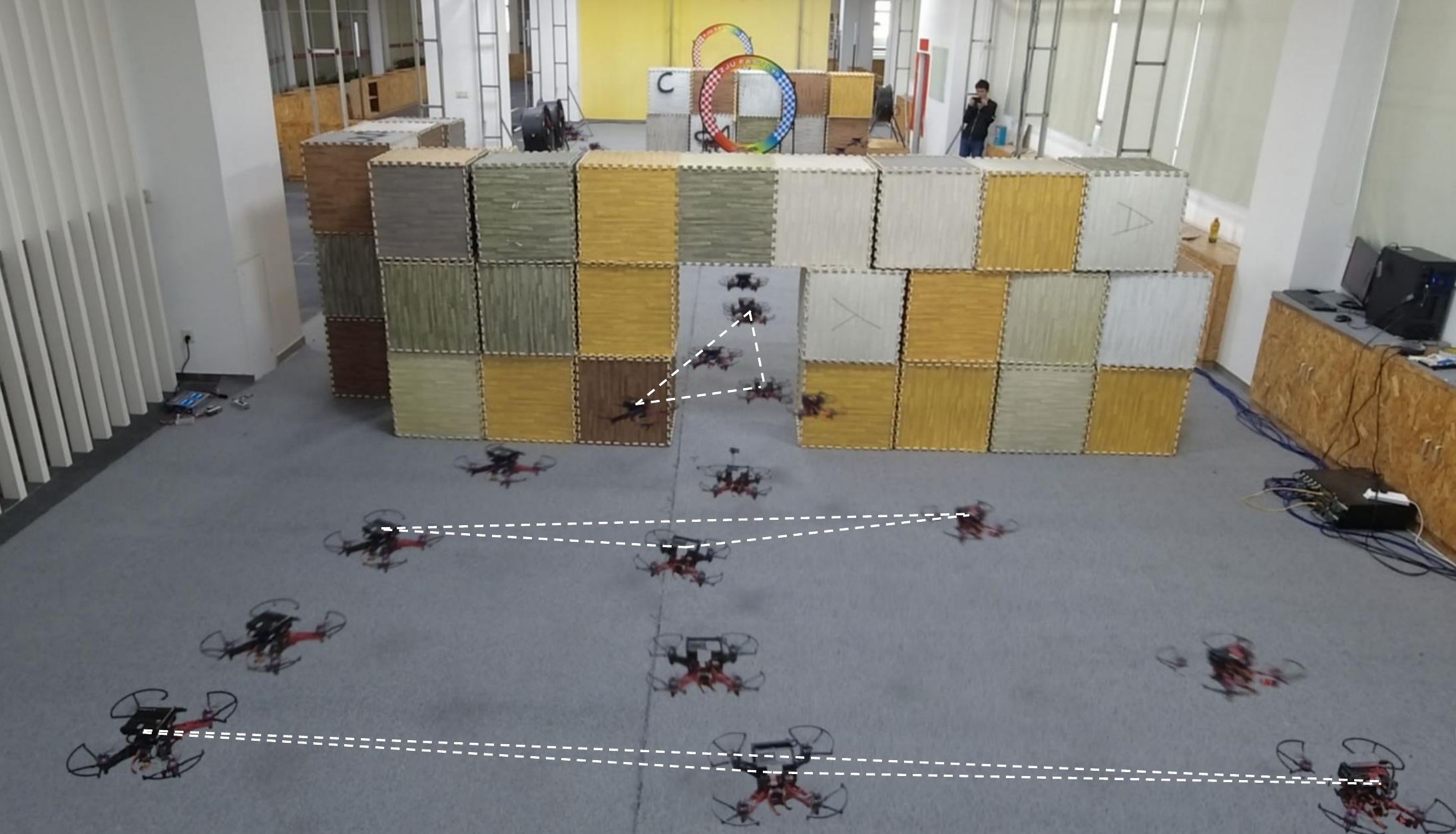}
			\captionsetup{font={small}}
			\caption{Narrow Gate}
			\label{pic:indoorgate}
		\end{subfigure}
		\begin{subfigure}{0.3\linewidth}
			\includegraphics[width=1\linewidth]{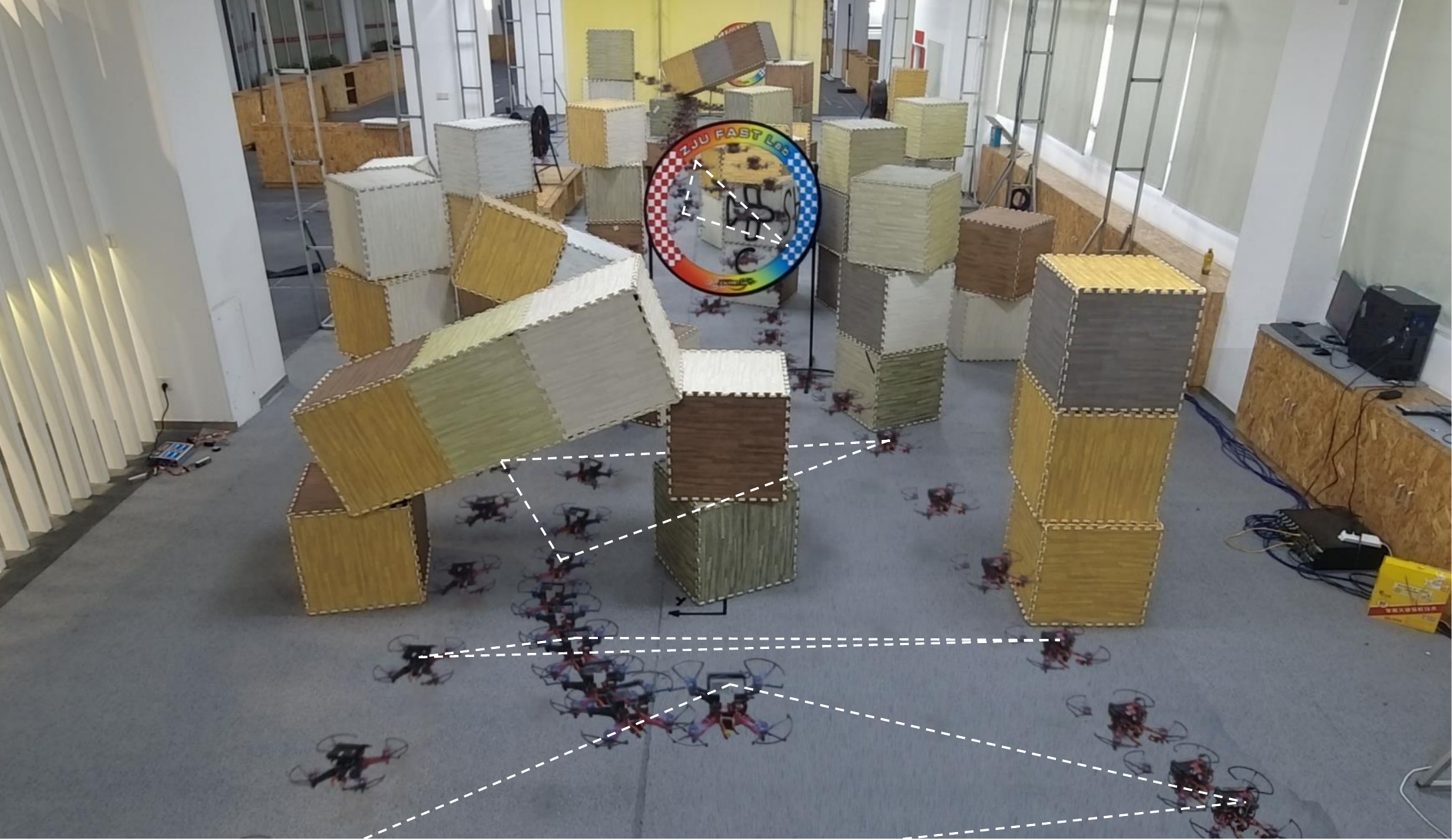}
			\captionsetup{font={small}}
			\caption{Dense Environment}
			\label{pic:indoorobs}
		\end{subfigure}
		\begin{subfigure}{0.91\linewidth}
			\includegraphics[width=1\linewidth]{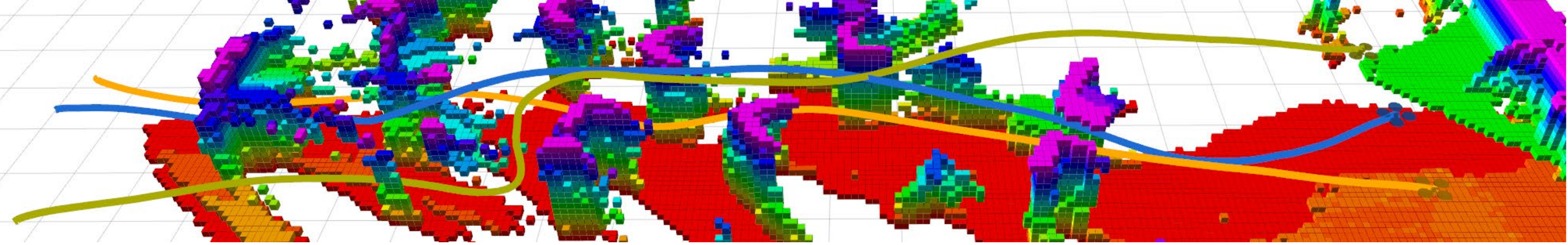}
			\captionsetup{font={small}}
			\caption{Trajectory and map created during the flight in dense environment.}
			\label{pic:indoormap}
		\end{subfigure}
		\captionsetup{font={small}}
		\caption{ Composite images of indoor experiments, in which three drones fly to three reverse placed targets. Dashed lines indicate the relative positions in the same photographs. As shown in the images, agents maintain a safe distance from each other while avoiding obstacles. Note that the trajectories are smooth, and there is no detour. }
		\label{pic:indoorexp}
	\end{figure*}
	
	\section{Real-World Experiments}
	\label{sec:experiments}
	
	\subsection{System Architecture}
	System architecture is depicted in Fig. \ref{pic:sys_archi}.
	A broadcast network that shares trajectories and performs on-demand time synchronization is the only connection among all agents.
	Therefore it is less coupled than \cite{zhou2020egoswarm}.
	
	In Fig. \ref{pic:sys_archi}, three of the modules may be confusing, here is a simple explanation.
	The module "Drone Detection" is for detecting other agents witnessed.
	The module "Frame Alignment" compensates single agent localization drift using the position deviation acquired from "Drone Detection" module.
	The module "Agent Removal" removes pixels of other agents from depth images, since these pixels can interfere the depth-based obstacle mapping.
	A detailed explanation of these three modules is in our previous work \cite{zhou2020egoswarm}.
	
	\subsection{Indoor}
	We present several indoor experiments at a speed limit of $3.0m/s$ for empty space and $2.0m/s$ for narrow gate and obstacle-rich scenarios, as depicted in Fig.\ref{pic:indoorexp}.
	The left one shows three quadrotors perform a cross flight in empty space where reciprocal collision avoidance is necessary.
	In the middle one, quadrotors manage to pass through a narrow gate one after another.
	In the right figure, we set up a cluttered environment composed of vertical and horizontal obstacles, where the narrowest gate is less than $1m$.
	In such a cluttered environment, three quadrotors manage to navigate across this environment sequentially and smoothly.
	
	\subsection{Outdoor}
	Outdoor experiments are presented to validate that the proposed system is capable of field operations.
	Snapshots of this experiment with the map built during the flight are shown in Fig. \ref{pic:head}.
	Please watch the video at \href{https://www.youtube.com/watch?v=w5GDMpjAoVQ}{https://www.youtube.com/watch?v=w5GDMpjAoVQ} for more information.
	The hardware settings are the same as \cite{gao2020teach}.

	\section{Conclusion and Future Work}
	\label{sec:conclusion}
	
	In this work, we propose a decentralized spatial-temporal trajectory planning framework for multicopter swarms.
	Our framework is powered by $\mathfrak{T}_{\rm{MINCO}}$ trajectory class capable of spatial-temporal deformation, and the time integral penalty functional based on constraint transcription.
	All these components enjoy efficiency originating from low-complexity.
	Extensive benchmarks are performed to show the speedup by orders of magnitude and the top-level solution quality.
	Real-word experiments also demonstrate the wide applicability of our framework.
	
	It's worth noting that the formulation of reciprocal collision avoidance in Sec. \ref{sec:swarm_avoidance} enables the planner to avoid moving obstacles as well.
	Since we use non-cooperative swarms, other agents and moving obstacles behave identically from the planner's view.
	However, we exclude this module in the current work due to the immature front-end of moving object detection and prediction, which is taken as the future work to enable fully autonomous aerial swarms in dynamical environments.

	\setlength{\bibsep}{-0.2ex}
	\bibliographystyle{plainnat}
	\bibliography{references}
	
\end{document}